\documentclass[pdflatex,sn-mathphys-ay]{sn-jnl}

\usepackage{graphicx}%
\usepackage{booktabs}%
\usepackage{multirow}%
\usepackage{amsmath,amssymb,amsfonts}%
\usepackage{amsthm}%
\usepackage{mathrsfs}%
\usepackage[title]{appendix}%
\usepackage{xcolor}%
\usepackage{textcomp}%
\usepackage{manyfoot}%
\usepackage{booktabs}%
\usepackage{algorithm}%
\usepackage{algorithmicx}%
\usepackage{algpseudocode}%
\usepackage{listings}%
\usepackage{subcaption}
\usepackage{array}
\usepackage{makecell}
\usepackage{anyfontsize}
\usepackage{url}
\usepackage{tabularx}
\usepackage[dvipsnames]{xcolor}

\theoremstyle{thmstyleone}%
\theoremstyle{thmstyletwo}%

\theoremstyle{thmstylethree}%

\begin{document}

\title[When, How Long and How Much? Interpretable Neural Networks for Time Series Regression by Learning to Mask and Aggregate]{When, How Long and How Much? Interpretable Neural Networks for Time Series Regression by Learning to Mask and Aggregate}


\author[1]{\fnm{Florent} \sur{Forest}$\textsuperscript{\orcid{https://orcid.org/0000-0001-6878-8752}}$}\email{f@florentfo.rest}
\equalcont{These authors contributed equally to this work.}

\author[1]{\fnm{Amaury} \sur{Wei}$\textsuperscript{\orcid{https://orcid.org/0000-0002-8626-9128}}$}\email{amaury.wei@epfl.ch}
\equalcont{These authors contributed equally to this work.}

\author*[1]{\fnm{Olga} \sur{Fink}$\textsuperscript{\orcid{https://orcid.org/0000-0002-9546-1488}}$}\email{olga.fink@epfl.ch}

\affil[1]{\orgdiv{IMOS Laboratory}, \orgname{EPFL}, \orgaddress{\city{Lausanne}, \postcode{CH-1015}, \country{Switzerland}}}


\abstract{Time series extrinsic regression (TSER) aims to predict a continuous target variable from an input time series. It arises in domains such as healthcare, finance, environmental monitoring, and engineering, where both accuracy and interpretability are essential. Despite strong predictive performance, state-of-the-art TSER models typically operate as black boxes, making it difficult to understand which temporal patterns drive their predictions. Post-hoc explanation methods attempt to address this issue but often produce coarse, noisy, or unstable explanations. Inherently interpretable approaches based on additive decompositions or concepts offer an alternative, yet they often require concept supervision, struggle to capture multivariate interactions, and lack expressiveness for complex temporal patterns.

To address these limitations, we propose MAGNETS (Mask-and-Aggregate Networks for Time Series), an inherently interpretable neural architecture for TSER. MAGNETS learns a compact set of human-understandable concepts directly from data, without requiring concept annotations. Each concept corresponds to a learned, mask-based aggregation over selected input features, revealing both \textit{which} features influence the prediction and \textit{when} they matter. Predictions are expressed as transparent combinations of these concepts, yielding faithful, input-specific explanations by design.

Experiments on synthetic and real-world univariate and multivariate TSER datasets show that MAGNETS achieves accuracy comparable to black-box models while substantially outperforming existing interpretable baselines, particularly on tasks involving multivariate feature interactions.
}

\keywords{Time series regression, Explainability, Concept learning, Interpretability}



\maketitle

\section{Introduction}\label{sec1}

Time series extrinsic regression (TSER) refers to the task of predicting a continuous target variable from an input time series \citep{tan_time_2021}. It appears in many domains\textemdash including healthcare (e.g., vital-sign forecasting), finance (e.g., volatility prediction), engineering (e.g., predicting battery state-of-charge or estimating remaining useful life from sensor streams), and environmental monitoring (e.g., pollution estimation). In these settings, accurate predictions and trustworthy reasoning are both essential \citep{mohammadi_foumani_deep_2024}. However, despite recent gains in accuracy, the opacity of modern TSER models remains an obstacle to adoption, particularly when predictions must be understood and validated by domain experts.

The demand for interpretability becomes especially difficult to satisfy with current state-of-the-art TSER models. Although recent approaches achieve strong predictive performance, they typically operate as black boxes. Leading approaches include deep neural networks \citep{ismail_fawaz_deep_2019,chen2024addressing}, large ensembles \citep{guijo-rubio_unsupervised_2024}, and RandOm Convolutional KErnel Transform (ROCKET) models \citep{dempster_rocket_2020,tan_multirocket_2022,dempster2023hydra}. Their complexity, large parameter counts, and lack of transparency make it difficult to understand which temporal features drive predictions or how multivariate interactions across variables influence the model’s output. These limitations hinder deployment in sensitive or regulated contexts. For example, if a TSER model predicts an impending system failure, engineers need to trace that prediction back to specific sensor behaviors and time intervals, rather than relying on an alert that cannot be meaningfully explained or validated.

These interpretability challenges have motivated growing interest in Explainable AI (XAI) research for time series, which aims to bridge the gap between predictive accuracy and actionable insight. While this research direction has made meaningful progress, existing approaches still fall short in several ways.
Post-hoc explanation methods, such as saliency maps and feature attribution \citep{ismail_benchmarking_2020,liu_timex_2024,jang2025timing}, attempt to rationalize a model’s decisions after training, but their explanations are often coarse, unstable, or poorly aligned with the model’s true internal reasoning \citep{rudin_stop_2019}.

In contrast, inherently interpretable models embed transparency directly into their architecture. Neural Additive Models (NAMs) \citep{agarwal_neural_2021} and their time-series extension, Neural Additive Time-series Models (NATMs) \citep{jo_neural_2023}, offer interpretable decompositions but remain fundamentally univariate, making it difficult for them to capture interactions between variables.
Concept-based approaches, such as Concept Bottleneck Models (CBMs) \citep{koh_concept_2020}, offer richer interpretability but require concept supervision and have only recently begun to be explored for time series \citep{forest_interpretable_2024}. Efforts in automated concept discovery for vision and language \citep{ghorbani_towards_2019,oikarinen_label-free_2023,rao_discover-then-name_2024,ludan_interpretable-by-design_2024} rely heavily on pretrained foundation models. Although time series foundation models have begun to emerge, they are primarily optimized for forecasting rather than regression, and the landscape remains fragmented, with no dominant architectures comparable to those in vision or natural language.
Recent efforts toward unsupervised concept discovery in time series \citep{wu_discovering_2022,alvarez-rodriguez_interpretable_2024} instead rely on global, input-independent statistical summaries. This design limits expressiveness and prevents them from capturing localized temporal patterns or multivariate interactions.

Consequently, no existing approach simultaneously provides (i) inherently interpretable predictions, (ii) annotation-free concept discovery, (iii) temporal localization, and (iv) ability to capture multivariate interactions.

To overcome these limitations, we propose MAGNETS (Mask-and-AGgregate NEtworks for Time Series), an inherently interpretable neural architecture for TSER.  MAGNETS learns input-dependent masks that identify locally relevant regions of each time series and aggregate these resulting masked segments into a compact set of predictive concepts, \textcolor{black}{targeting regression tasks driven by discrete, localized temporal events.}

Unlike previous concept-based methods, MAGNETS does not rely on predefined concept annotations or pretrained foundation models. Instead, it discovers a compact set of human-understandable concepts directly from raw inputs, without any concept supervision. Each concept corresponds to a localized, mask-based aggregation over selected variables, jointly revealing \textit{which} variables matter and \textit{when} they influence the prediction. Predictions are then constructed as combinations of these learned concepts, providing a transparent and structured reasoning process that users can inspect and visualize.

We evaluate MAGNETS on synthetic and real-world TSER benchmarks covering both univariate and multivariate settings. Across all datasets, MAGNETS achieves predictive accuracy competitive with state-of-the-art black-box models while offering substantially richer interpretability. It consistently outperforms existing interpretable baselines, particularly in tasks involving variable interactions or localized temporal patterns. Its explanations are also more faithful and informative than those produced by post-hoc attribution methods. \textcolor{black}{Overall, MAGNETS directly addresses the performance-interpretability trade-off, narrowing the gap with black-box baselines while outperforming existing interpretable architectures.}

The main contributions of this work are as follows:
\begin{itemize}
    \item We introduce MAGNETS, a new interpretable neural architecture for time series extrinsic regression that learns localized, mask-based concept aggregations.
    \item We propose an annotation-free concept bottleneck mechanism that discovers meaningful concepts capable of capturing both multivariate interactions and temporal localization, capabilities that lie beyond existing transparent models.
    \item We conduct extensive qualitative and quantitative evaluations demonstrating that MAGNETS delivers competitive predictive accuracy while providing superior interpretability and richer insights compared to post-hoc XAI methods.
\end{itemize}

The remainder of this paper is organized as follows. Section \ref{sec:related} reviews related work on interpretable modeling and XAI for time series. Section \ref{sec:method} presents the MAGNETS architecture and training procedure. Section \ref{sec:experiments} details the datasets, and experimental setup. Section \ref{sec:results} reports the results and discussion. Section \ref{sec:conclusion} presents the conclusions and outlines directions for future research.

\section{Related Work}
\label{sec:related}

Research on interpretability for time series models falls into three main categories: (i) post-hoc explanation methods, (ii) inherently interpretable architectures, and (iii) concept-based modeling. We review each of these directions below, highlighting their respective strengths and limitations in the context of time series extrinsic regression.

\subsection{Post-hoc XAI for Time Series}
Post-hoc interpretability methods aim to explain already-trained models by estimating the contribution of individual input features or specific regions of the time series \citep{arrieta_explainable_2019}. Classical attribution methods such as Integrated Gradients (IG) \citep{sundararajan_axiomatic_2017} and DeepLIFT (DL) \citep{shrikumar_learning_2019} are widely used in vision and tabular domains. However, extending these methods to time series is challenging due to strong autocorrelations and complex temporal dependencies \citep{ismail_benchmarking_2020}.

To address these challenges, several post-hoc methods tailored specifically to time series have been proposed. Feature Importance in Time (FIT) \citep{tonekaboni_what_2020} quantifies the incremental predictive value of each time step relative to its history, offering an improvement over naive perturbation-based approaches. DynaMask \citep{crabbe_explaining_2021} learns dynamic occlusion masks under sparsity and smoothness constraints using differentiable perturbations. WinIT \citep{leung_temporal_2023} extends FIT by computing importance over multi-step sliding windows. Other recent methods further improve masking strategies by learning context-dependent perturbations \citep{enguehard_learning_2023} or by generating in-distribution masked samples, as in TimeX \citep{queen_encoding_2023} and TimeX++ \citep{liu_timex_2024}. Example-based explanations have also been explored, for instance by extracting representative subsequences from intermediate Convolutional Neural Network (CNN) activations \citep{younis_multivariate_2022,s21062154}.

While these approaches provide useful post-hoc insights, they share several fundamental limitations: their explanations are not guaranteed to reflect the model’s true internal reasoning, they may vary substantially across methods, and they impose no interpretability constraints during training. Moreover, most existing work focuses on classification tasks, leaving interpretability for TSER largely underexplored.

\subsection{Interpretable Models for Time Series}

Contrary to post-hoc approaches, inherently interpretable models enforce transparency through their architecture, ensuring that the computation pathway can be inspected directly.

\textbf{Additive Models.}\quad Generalized Additive Models (GAMs) \citep{hastie_generalized_1986} decompose predictions as a sum of univariate  shape functions, enabling users to inspect each feature's isolated contribution. While originally developed for tabular data, these models are not tailored to temporal structure. Neural extensions such as Generalized Additive Neural Networks (GANNs) \citep{potts_generalized_1999,kraus_interpretable_2024} and Neural Additive Models (NAMs) \citep{agarwal_neural_2021} replace shape functions with small neural networks, enhancing expressiveness while preserving additive decomposition. Multiple extensions attempting to add feature interactions exist \citep{vaughan_explainable_2018,yang_gami-net_2021,chang_node-gam_2022}, but they remain difficult to scale to higher-order or high-dimensional settings \citep{radenovic_neural_2022}.

Additionally, extending these inherently interpretable architectures to time series remains nontrivial. Time series introduce additional challenges due to temporal dynamics and multivariate interactions. Neural Additive Time-series Models (NATMs) \citep{jo_neural_2023} adapt NAMs to temporal data by learning a shape function per channel and time step, optionally sharing parameters across dimensions to reduce complexity. However, NATMs remain fundamentally univariate: they model each channel independently and therefore cannot capture cross-channel interactions. More recently, the Generalized Additive Time Series Model (GATSM) \citep{kim_transparent_2024} employs neural basis functions with causal attention to scale to longer sequences, but still inherits the additive restriction, limiting its ability to express multivariate temporal interactions. Although additive models offer strong interpretability, their limited expressiveness restricts their applicability in TSER tasks which involve interacting sensors or conditional temporal logic.

\textbf{Prototypes and Shapelets.}\quad Another family of interpretable models relies on comparing inputs to learned prototypes or discriminative subsequences. Originally developed for image classification, prototype networks \citep{li_deep_2017, gee_explaining_2019, ming_interpretable_2019, obermair_example_2023} explain predictions via similarity to representative latent patterns. Shapelet-based methods \citep{ye_time_2011} identify subsequences most predictive of the output, and gradient-based shapelet learning \citep{grabocka_learning_2014, medico_learning_2021} improves scalability by optimizing these subsequences directly. However, the learned shapelets may drift away from realistic or physically plausible patterns \citep{qu_cnn_2024}, and these approaches are primarily designed for classification. They typically rely on nearest-neighbor or similarity-based decision rules and do not naturally extend to regression without discretizing the target variable \citep{guijo-rubio_time_2020}. Consequently, their applicability to TSER remains limited.

\subsection{Concept-based Models}

Concept-based approaches seek to structure the prediction through higher-level, human-interpretable concepts rather than individual time points.

\textbf{Supervised Concept Bottlenecks.}\quad Self-Explaining Neural Networks (SENN) \citep{alvarez-melis_towards_2018} combine interpretable concept encoders with input-dependent relevance scores to produce transparent predictions. Concept Bottleneck Models (CBMs) \citep{koh_concept_2020} separate learning into a concept prediction stage followed by a label prediction stage based on those concepts. Concept Embedding Models (CEMs) \citep{zarlenga_cem_2022} extend CBMs by learning continuous concept embeddings instead of binary activations, improving accuracy at the expense of interpretability. A fundamental limitation of all these supervised approaches is their reliance on concept annotations, which are rarely available in most time series domains. As a result, applications to temporal data remain limited.  The only TSER application to date focuses on aircraft engine health prediction using manually curated degradation concepts \citep{forest_interpretable_2024}.

\textbf{Unsupervised Concept Discovery.}\quad Unsupervised concept discovery has seen major advances in vision and NLP, e.g. using superpixels (Automated Concept-based Explanation (ACE) by \cite{ghorbani_towards_2019}), label-free CBMs \citep{oikarinen_label-free_2023}, or multimodal supervision via foundation models \citep{tan_explain_2024, ludan_interpretable-by-design_2024}. However, these techniques rely on spatial structure or pretrained models not available for time series.

Only a few works explore unsupervised concept discovery in temporal data. Optimal Summaries (OS) \citep{wu_learning_2022} treat global statistical aggregations (e.g. mean, variance, slope, thresholded durations) as concepts and optimize their parameters within a CBM. However, these summaries are fixed across inputs, cannot localize temporal patterns, and remain univariate. A similar statistical-summary formulation is proposed in \citep{alvarez-rodriguez_interpretable_2024}. Shapelet bottlenecks with rule extraction \citep{wen_shedding_2024} combine concept-like shapelets with a black-box expert via a gating mechanism, but this approach remains classification-oriented. In time-series forecasting, surrogate concept tasks have been used to guide transformer training \citep{sprang_enforcing_2024}, but this formulation is specific to predicting future time steps and does not extend to TSER.

Overall, unsupervised, temporally localized, multivariate concept discovery for time series regression remains an open challenge, motivating the development of new approaches.

\section{Method}\label{sec:method}

\begin{figure*}
    \centering
    \includegraphics[width=\linewidth]{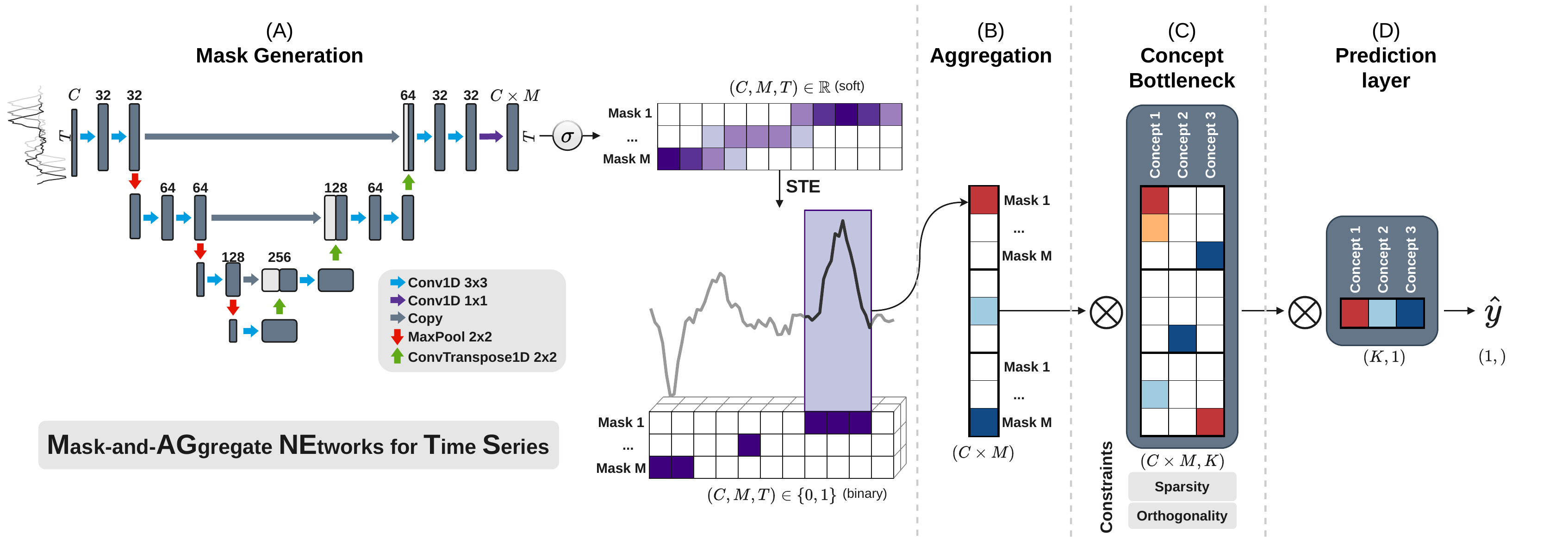}
    \caption{Overview of the proposed MAGNETS architecture for interpretable TSER. (A) Given an input multivariate time series, the mask generation network produces a set of logits (Eq.~\ref{eq:mask_gen_logits}), discretized into binary masks using the Gumbel softmax with Straight-Through Estimator (STE) (Eqs.~\ref{eq:mask_gen_gumbel}\textendash\ref{eq:mask_gen_ste}). (B) The selected temporal regions of the input are aggregated by integrating over the time dimension. (C) The aggregated features pass through a linear concept bottleneck layer with sparsity and orthogonality constraints. (D) Finally, a transparent linear prediction layer maps concept activations to the final output $\hat{y}$.}
    \label{fig:magnets-overview}
\end{figure*}

In this section, we describe the detailed architecture and components of the proposed MAGNETS model for interpretable TSER. Let $\mathbf{x}\in\mathbb{R}^{C\times T}$ denote a multivariate time series (MTS) with $C$ the number of channels and $T$ the number of time steps. The goal in TSER is to learn a predictor $f:\mathbb{R}^{C\times T}\to\mathbb{R}$ mapping $\mathbf{x}$ to a continuous target value $\hat{y} \in \mathbb{R}$.

To ensure transparency in its predictions, MAGNETS decomposes the regression task into a sequence of four simple and interpretable operations (illustrated in Figure~\ref{fig:magnets-overview}):

\begin{enumerate}[(A)]
    \item \textbf{Mask generation}. A neural network $\phi^{\mathrm{masks}}$ predicts $M$ masks per channel $\mathbf{m}^{\mathrm{soft}} = \phi^{\mathrm{masks}}(\mathbf{x}) \in \mathbb{R}^{C \times M \times T}$ which are binarized into masks $\mathbf{m} \in \{0,1\}^{C \times M \times T}$, selecting input-dependent temporal regions.
    \item \textbf{Aggregation}. Each mask  $\mathbf{m}_{c,m}$ is applied to the input by element-wise multiplication $\tilde{x}_{c,m,t} = x_{c,t} \cdot m_{c,m,t}$, and aggregated over time to produce scalar interpretable features $z_{c,m} = \sum_{t=1}^T \tilde{x}_{c,m,t}$.
    \item \textbf{Concept bottleneck.} A linear layer $\phi^{\mathrm{bottleneck}}$ maps the aggregated features $\mathbf{z} \in \mathbb{R}^{C\times M}$ to $K$ concept activations $\{c_k\}_{k=1}^K$ with sparsity and orthogonality regularizations for interpretability.
    \item \textbf{Prediction layer.} A final transparent linear layer $\phi^{\mathrm{prediction}}$ produces the final prediction $\hat{y}$ as a linear combination of the concept activations.
\end{enumerate}

\textcolor{black}{Designed for tasks driven by discrete, localized temporal events}, our pipeline answers two key interpretability questions about the input:

\begin{itemize}
    \item \textbf{When do relevant temporal patterns occur?} The mask generator identifies input-dependent temporal regions and outputs a set of binary masks per input channel.
    \item \textbf{How long and how much} do these patterns contribute? Aggregation over the selected input regions captures both their duration (how long) and their intensity (how much).
\end{itemize}

We describe each component in detail below.

\subsection{Mask Generation}

The mask generation network $\phi^{\mathrm{masks}}$ is responsible for identifying the temporal regions of the input time series that are informative for predicting the target value $\hat{y}$. For each input $\mathbf{x}\in\mathbb{R}^{C\times T}$, the network predicts  a tensor of mask logits
\begin{equation}\label{eq:mask_gen_logits}
\mathbf{m}^{\mathrm{soft}} = \phi^{\mathrm{masks}}(\mathbf{x}) \in \mathbb{R}^{C \times M \times T}
\end{equation}
where $m^{\mathrm{soft}}_{c,m,t}$ represents the pre-activation score for channel $c$, mask index $m$, and time step $t$. Masks are channel-specific, meaning each channel therefore receives $M$ mask candidates, allowing the model to capture multiple, possibly disjoint temporal patterns.

The architecture of $\phi^{\mathrm{masks}}$ is flexible and can be chosen according to the complexity of the task. Conceptually, the mask generator plays a role analogous to an image segmentation network \citep{stalder_what_2022}, identifying relevant regions of interest\textemdash here along the temporal axis rather than across a spatial grid.

In this work, we adopt a 1D U-Net model \citep{iglovikov_ternausnet_2018} as $\phi^{\mathrm{masks}}$. This encoder-decoder CNN with skip connections effectively captures both local and global temporal dependencies. Although $\phi^{\mathrm{masks}}$ is a high-capacity neural network, \textcolor{black}{it is restricted exclusively to proposing candidate temporal regions. The subsequent computational path operates directly on the raw data within these segments, ensuring a fully interpretable and transparent prediction.}

\textbf{\textcolor{black}{Binary} masks for continuous time series.}\quad As noted by~\cite{queen_encoding_2023}, \textcolor{black}{binary} (rather than soft or probabilistic) masks are preferable for interpretability in continuous time series. In discrete token modalities such as text or graphs\textemdash where soft masking can still preserve semantic meaning\textemdash this distinction is less critical. In contrast, time series regression depends on the precise magnitude and shape of the input values. Soft masks would blur temporal regions, obscuring which parts of the signal influence the prediction and thereby weakening interpretability.

To enable gradient-based training despite the binary nature of the masks, we apply the Straight-Through Gumbel-Softmax estimator \citep{jang_categorical_2017}, adopting its binary (logistic) variant as in \citep{liu_timex_2024}. For each logit $m^{\mathrm{soft}}_{c,m,t}$, we construct a relaxed binary mask value
\begin{equation}\label{eq:mask_gen_gumbel}
\tilde{m}_{c,m,t} = \sigma\!\left(\frac{m^{\mathrm{soft}}_{c,m,t} + g_{c,m,t}}{\tau}\right),
\end{equation}
where $\sigma(\cdot)$ is the logistic sigmoid, $g_{c,m,t}$ is i.i.d.\ Gumbel noise, and $\tau>0$ is a temperature that controls the sharpness of the relaxation. During the forward pass, this value is binarized as
\begin{equation}
m_{c,m,t} = \mathbb{I}\!\left(\tilde{m}_{c,m,t} > 0.5\right),
\end{equation}
yielding a binary mask. During the backward pass, however, gradients are propagated through $\tilde{m}_{c,m,t}$:
\begin{equation}\label{eq:mask_gen_ste}
\frac{\partial m_{c,m,t}}{\partial s_{c,m,t}}
\;\approx\;
\frac{\partial \tilde{m}_{c,m,t}}{\partial s_{c,m,t}}.
\end{equation}
This straight-through estimator (STE) maintains stable gradient flow throughout training while ensuring that masks are strictly binary  at inference time. The resulting tensor
\begin{equation}
\mathbf{m} \in \{0,1\}^{C \times M \times T}
\end{equation}
specifies, for every channel and mask, which time steps are selected. These masks are input-specific\textemdash rather than fixed global patterns\textemdash and explicitly encode “\emph{when}” the model attends to the signal in the subsequent aggregation stage. 

\subsection{Aggregation}

Once the binary masks are obtained, they are applied to the input time series to isolate the temporal regions identified as relevant for prediction. For each channel $c$ and mask index $m$, the masked signal is defined element-wise as
\begin{equation}
\tilde{x}_{c,m,t} = x_{c,t} \cdot m_{c,m,t},
\end{equation}
which preserves the original values at selected time steps while zeroing out all others. Because mask values are produced independently at each time index, a single mask may select one or several disjoint temporal intervals, allowing the model to focus adaptively on multiple informative segments within a channel. All masks are input-specific rather than global, which enables instance-level interpretability.

To convert each masked segment into a meaningful feature, MAGNETS applies a user-specified aggregation function
\begin{equation}\label{eq:aggregation}
    g:\mathbb{R}^{T}\to\mathbb{R}
\end{equation}
which aggregates the masked values $\tilde{\mathbf{x}}_{c,m}=(\tilde{x}_{c,m,1},\ldots,\tilde{x}_{c,m,T})$ into a single scalar. Using scalar summaries provides a direct, one-to-one correspondence between each channel-mask pair and its numerical contribution to the model, making the influence of each masked region easy to inspect and compare. Higher-dimensional aggregations would dilute this one-to-one correspondence and obscure how masks influence the final prediction.

The aggregation function $g$ is specified explicitly rather than learned. This avoids introducing hidden nonlinear transformations whose semantics may be difficult to interpret, and allows practitioners to choose summaries  aligned with domain knowledge. The only requirement is differentiability, enabling choices such as the mean, an $\ell_p$-norm, or other smooth statistics depending on the task and the temporal or cross-channel dependencies present in the data. In this work, we adopt the summation operator:
\begin{equation}
z_{c,m} = \sum_{t=1}^{T} \tilde{x}_{c,m,t}
\end{equation}

Summation offers a particularly interpretable quantity: it simultaneously reflects the duration of the selected region (how long it is active) and the magnitude of the underlying signal (how strongly it contributes). This cumulative measure aligns naturally with many real-world TSER settings\textemdash for example, the total time a sensor exceeds a safety threshold or the integrated energy output of an engineering system\textemdash making it a robust and intuitive choice across applications.

The aggregation step produces a set of $C\times M$ scalar features, one per channel–mask pair, enabling MAGNETS to capture a rich set of localized, instance-specific temporal effects. This stands in contrast to approaches such as Optimal Summaries \citep{wu_learning_2022}, which rely on predetermined global cutoffs and therefore cannot adapt their summaries to the  structure of each individual input series.

\textcolor{black}{\textbf{Relationship to attention mechanisms.}\quad Conceptually, MAGNETS differs from standard attention~\citep{vaswani2017attention} in two ways to maximize interpretability. First, it replaces continuous softmax weights with strict $\{0,1\}$ binary selection (via the Straight-Through estimator) to extract crisp temporal intervals. Second, it bypasses learned Query-Key-Value (QKV) projections, instead applying masks directly to the raw input with a parameter-free summation readout. By preserving physical signal magnitudes, these modifications ensure the extracted features transparently answer \emph{when}, \emph{how long}, and \emph{how much} a pattern contributes.}

\subsection{Concept Bottleneck}

The purpose of the concept bottleneck $\phi^{\mathrm{bottleneck}}$ is to map the aggregated, mask-specific features $\{z_{c,m}\}$ into a smaller set of higher-level, human-interpretable concepts. Unlike supervised concept bottleneck models \citep{koh_concept_2020}, MAGNETS discovers these concepts automatically during training, removing the need for concept annotations. Each concept emerges as a learned linear combination of the scalar features derived from masked temporal regions. Let
\begin{equation} \label{eq:agg_feature_dual}
\mathbf{z} = \{z_{c,m}\}_{c=1, m=1}^{C,\quad M} \in \mathbb{R}^{C\times M}
\end{equation}
denote the aggregated features. The bottleneck applies a linear transformation:
\begin{equation} \label{eq:cbl_linear_dual}
    c_k = \sum_{c=1}^C \sum_{m=1}^M \beta_{c,m,k} \cdot z_{c,m} + b_k
\end{equation}
where $\mathbf{\beta} \in \mathbb{R}^{C \times M \times K}$ contains the bottleneck weights and $b_k$ the bias associated with concept $k$. Because the transformation is linear and operates directly on the scalar aggregated features, the contribution of each masked region to each concept remains explicit and easy to interpret.

\medskip
The bottleneck contains $C \times M \times K$ parameters, which may lead to dense or redundant concepts if left unconstrained. To keep the concepts both compact (each depending on only a few features) and non-redundant (each capturing different temporal patterns), we introduce two complementary regularization terms.

\textbf{Sparsity loss.}\quad To encourage each concept to rely on only a small subset of aggregated features, we apply an $\ell_1$ penalty:
\begin{equation}
\mathcal{L}_{\mathrm{spars}}(\boldsymbol{\beta}) = |\mathbf{\boldsymbol{\beta}}|_1.
\end{equation}
which encourages most weights to shrink toward zero.
This promotes concepts that depend on a limited number of channel–mask pairs, reducing entanglement and improving interpretability, consistent with findings in transparent TSER models \citep{wu_learning_2022}.

\textbf{Orthogonality loss.}\quad To reduce redundancy between concepts\textemdash i.e. to prevent different concepts from encoding similar combinations of features\textemdash we encourage their weight vectors to be approximately orthogonal:
\begin{equation}
\mathcal{L}_{\text{ortho}}(\boldsymbol{\beta}) = |\boldsymbol{\beta}^\top \boldsymbol{\beta} - \mathbf{I}|_F^2,
\end{equation}
where $\mathbf{I}$ is the identity matrix and $|\cdot|_F$ denotes the Frobenius norm. While this does not enforce perfect orthogonality, it meaningfully limits overlap in how concepts combine the aggregated features, thereby promoting diversity among the learned explanations.

\textbf{Resulting interpretability benefits.}\quad Together, sparsity and orthogonality produce concepts that are both compact and distinct, making it clear which masked temporal regions\textemdash and which channels\textemdash drive each concept activation.

\subsection{Prediction Layer}

The final stage of MAGNETS maps the concept activations to the scalar prediction $\hat{y}$. To preserve full transparency, this mapping is kept strictly linear:
\begin{equation}
\hat{y} = \phi^{\mathrm{prediction}}(\mathbf{c})= \sum_{k=1}^K w_k \cdot c_k + w_0
\end{equation}
where $w_k$ are the weights of $\phi^{\mathrm{prediction}}$ and $w_0$ is a bias term.

The linearity of $\phi^{\mathrm{prediction}}$ ensures that each concept contributes additively and with an explicit weight to the final output. Combined with the linear structure of the concept bottleneck, this makes every prediction directly traceable to specific parts of the input $\mathbf{x}$.

\subsection{Loss Function}

Training MAGNETS requires balancing predictive accuracy with the interpretability of the learned concepts. The overall objective $\mathcal{L}_{\mathrm{MAGNETS}}(\mathbf{x}, y)$ is defined as:
\begin{equation}
\mathcal{L}_{\mathrm{MSE}}(y, \hat{y}) + \lambda_{\mathrm{spars}} \mathcal{L}_{\mathrm{spars}}(\boldsymbol{\beta}) + \lambda_{\mathrm{ortho}} \mathcal{L}_{\mathrm{ortho}}(\boldsymbol{\beta})
\end{equation}
where $\mathcal{L}_{\mathrm{MSE}}$ encourages accurate regression, and the positive coefficients $\lambda_{\mathrm{spars}}$ and $\lambda_{\mathrm{ortho}}$ control the strength of the sparsity and orthogonality regularizers, respectively.

This objective is fully compatible with gradient-based optimization and encourages MAGNETS to achieve accurate predictions while preserving a clear, decomposable pathway, from masked temporal regions, to aggregated features, to concepts, and ultimately to the output.

\section{Experiments}\label{sec:experiments}

We evaluate MAGNETS on a suite of synthetic and real-world TSER datasets along three dimensions: 
(i) predictive accuracy relative to black-box and interpretable baselines, (ii) interpretability, measured by the model’s ability to recover ground-truth temporal logic, and  (iii) explanation quality compared to state-of-the-art post-hoc XAI methods.

Our evaluation covers four synthetic datasets\textemdash designed as controlled, explainability-focused benchmarks\textemdash and four real-world datasets from energy, environmental, and structural monitoring domains. All code and data are publicly available.

\subsection{Compared Methods}

\textbf{Black-box baselines.}\quad We compare MAGNETS to widely used high-performing TSER models: Random Forests \citep{breiman2001random}, ROCKET \citep{dempster_rocket_2020}, MultiROCKET \citep{tan_multirocket_2022}, ResNet~\citep{7966039}, InceptionTime~\citep{ismail2020inceptiontime}, and a CNN whose encoder is matched in capacity to the masking network used in MAGNETS. Exact architectural details are provided in Appendix~\ref{sec:appendix_hyperparameters}.

\textbf{Interpretable baselines.}\quad We include eight transparent or structure-aware models: (i) linear, Lasso, and Ridge regression; (ii) the three NATM variants \citep{jo_neural_2023}, reimplemented due to absence of released code; (iii) GATSM \citep{kim_transparent_2024} using the official implementation; (iv) and Optimal Summaries (OS) \citep{wu_learning_2022}.  
Because OS was originally designed for logistic classification, we apply three light modifications to ensure fairness in a regression setting: (a) replacing the logistic head with a linear regressor, (b) removing measurement-indicator features not present in our datasets, and (c) adapting threshold-based features to use “aggregated-above-threshold" values consistent with our aggregation scheme.

This selection spans linear, additive, kernel-based, summary-based, and deep learning paradigms, providing a broad and representative comparison against MAGNETS.

\subsection{Synthetic Datasets}

\begin{figure}[ht]
    \centering
    \begin{subfigure}[b]{0.45\linewidth}
        \includegraphics[width=\textwidth]{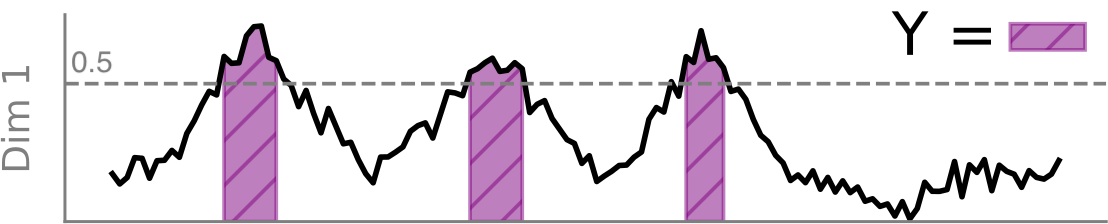}
        \caption{Univariate}
        \label{fig:synth:univariate}
    \end{subfigure}
    \begin{subfigure}[b]{0.45\linewidth}
        \includegraphics[width=\textwidth]{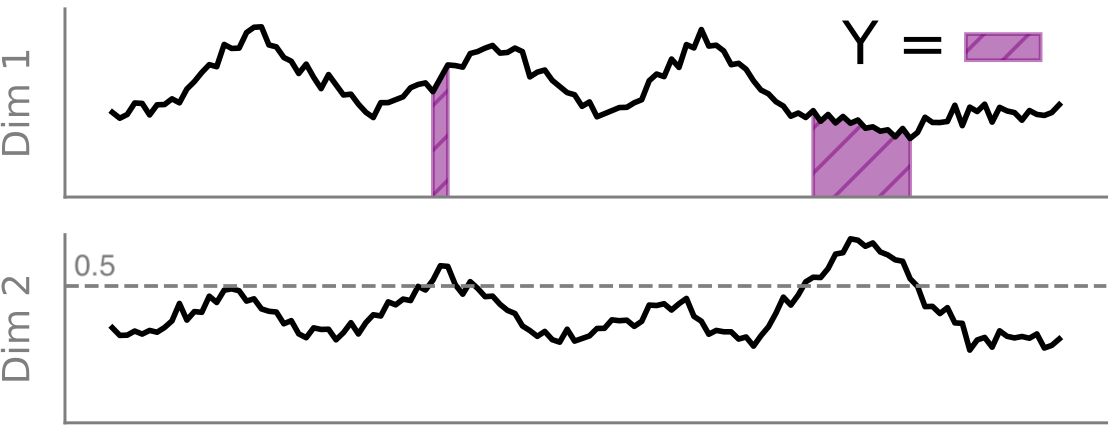}
        \caption{Bivariate}
        \label{fig:synth:bivariate}
    \end{subfigure}\\
    \begin{subfigure}[b]{0.45\linewidth}
        \includegraphics[width=\textwidth]{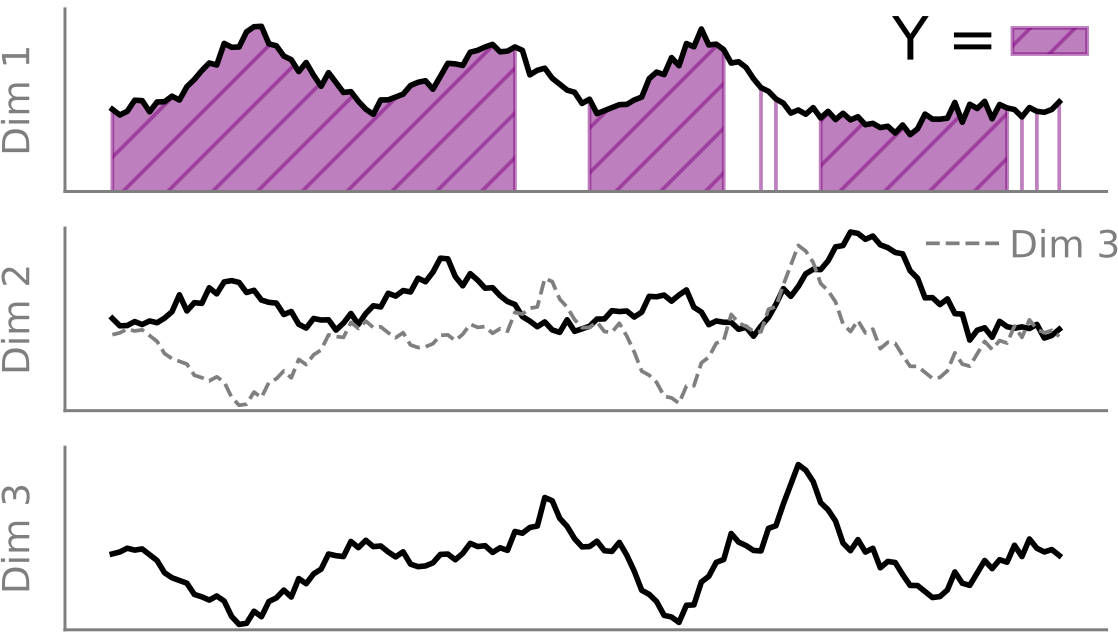}
        \caption{Trivariate-1}
        \label{fig:synth:trivariate-1}
    \end{subfigure}
    \begin{subfigure}[b]{0.45\linewidth}
        \includegraphics[width=\textwidth]{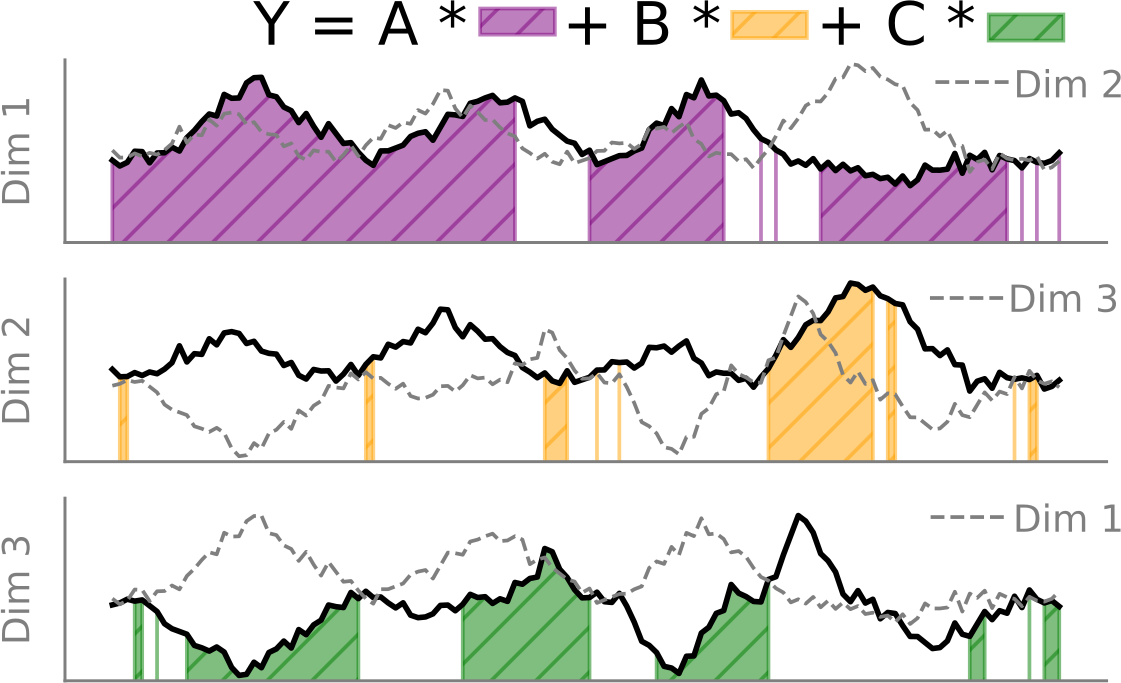}
        \caption{Trivariate-2}
        \label{fig:synth:trivariate-2}
    \end{subfigure}
    \caption{Illustrative examples from the four synthetic TSER datasets. Regression targets are defined as areas under specific regions: above a fixed threshold (Univariate, Bivariate) or satisfying channel-wise inequalities (Trivariate-1, Trivariate-2). The Trivariate-2 target is a weighted sum of three conditional areas with coefficients $(A,B,C)=(1,5,-2)$.}
    \label{fig:synth}
\end{figure}

\textcolor{black}{Although interpretability has become increasingly critical in TSER, real-world open-source datasets with ground-truth temporal annotations remain unavailable (to the best of our knowledge). While manual labeling is technically possible, it requires extensive domain expertise and is prohibitively expensive. Therefore, to quantitatively evaluate the ability to recover temporal patterns and channel interactions,} we construct four synthetic datasets with controlled logic. This enables rigorous assessment of whether models can recover the correct temporal dependencies and multivariate interactions. Dataset properties are summarized in Table~\ref{tab:datasets} and example series are shown in Fig.~\ref{fig:synth}.

\begin{itemize}
    \item \textbf{Univariate}: the target $y$ equals the area under the curve when the signal exceeds a threshold of 0.5.
    \item \textbf{Bivariate}: the target $y$ equals the area under channel 1 when channel 2 exceeds 0.5.
    \item \textbf{Trivariate-1}: the target $y$ equals the area under channel 1 when channel 2 is greater than channel 3.
    \item \textbf{Trivariate-2}: the target $y$ equals the weighted sum of three conditional areas defined by pairwise channel comparisons.
\end{itemize}

These synthetic datasets evaluate a model’s ability to localize temporal patterns, capture multivariate dependencies, and recover the  underlying explanatory logic.

\subsection{Real-world Datasets}

We further evaluate MAGNETS on ten real-world TSER datasets spanning energy consumption, wind power forecasting, air-quality monitoring, environmental risk modeling, and engineering asset health. Eight of them originate from the TSER archive \citep{tan_time_2021, guijo-rubio_unsupervised_2024}. \textcolor{black}{As MAGNETS is inherently designed for tasks driven by localized temporal aggregations, we explicitly selected datasets whose physical interpretation is plausibly aligned with localized accumulation, event duration, or interval-specific temporal effects. Datasets relying primarily on global sequence-level properties (\textit{e.g.}, spectroscopy or text embeddings) were excluded, as such modalities are poorly aligned with MAGNETS' localized mask-and-aggregate inductive bias. Alongside these archive datasets, we use the BatteryDegradation1 dataset~\citep{Birkl2017OxfordBD} and release a novel BridgeDegradation dataset~\citep{zhao2025disentanglingslowfasttemporal}.} Table~\ref{tab:datasets} summarizes their sample counts, lengths, and dimensionalities.

\begin{itemize}
    \item \textbf{HouseholdPowerConsumption1}: predict total daily active power from minute-level electrical measurements.  
    \item \textbf{WindTurbinePower}: predict daily wind turbine output from torque measurements.  
    \item \textbf{BenzeneConcentration}: predict hourly benzene concentration from chemical and environmental sensor readings.
    \item \textbf{BeijingPM10Quality \& BeijingPM25Quality}: predict pollutant concentration levels from multivariate meteorological and air-quality sensor streams.
    \item \textbf{FloodModeling (1, 2, 3)}: predict peak water height of riverbeds from a sequence of rainfall events.
    \item \textbf{BridgeDegradation}: predict bridge structural damage during a loading cycle from mid-span displacement, ambient temperature, and train load signals. Adapted from~\cite{zhao2025disentanglingslowfasttemporal}.
    \item \textbf{BatteryDegradation1}: predict a continuous state-of-health degradation metric from active discharge cycle sensors (voltage, current, and temperature).
\end{itemize}

These datasets cover diverse application domains, sequence lengths, channel counts,  noise characteristics, and degrees of nonstationarity, providing a comprehensive evaluation of MAGNETS under varied real-world conditions.

\begin{table}[h]
    \caption{Time series regression dataset properties: type, name, length $T$, dimensions $C$, and training/testing sizes.}
    \label{tab:datasets}
    \centering
    \setlength{\tabcolsep}{4pt}
    \begin{tabular}{llcccc}
        \toprule
        Type & Dataset name & $T$ & $C$ & $N_{\mathrm{train}}$ & $N_{\mathrm{test}}$ \\
        \midrule
        \multirow{4}{*}{Synthetic} & Univariate & 128 & 1 & 50'000 & 10'000 \\
        & Bivariate & 128 & 2 & 50'000 & 10'000 \\
        & Trivariate-1 & 128 & 3 & 50'000 & 10'000 \\
        & Trivariate-2 & 128 & 3 & 50'000 & 10'000 \\
        \midrule
        \multirow{10}{*}{Real} & HouseholdPowerConsumption1 & 144 & 5 & 745 & 686 \\
        & WindTurbinePower & 144 & 1 & 441 & 196 \\
        & BenzeneConcentration & 240 & 8 & 3'349 & 5'163 \\
        & BeijingPM10Quality & 24 & 9 & 12'432 & 5'100 \\
        & BeijingPM25Quality & 24 & 9 & 12'432 & 5'100 \\
        & FloodModeling1 & 266 & 1 & 471 & 202 \\
        & FloodModeling2 & 266 & 1 & 389 & 167 \\
        & FloodModeling3 & 266 & 1 & 429 & 184 \\
        & BridgeDegradation & 144 & 3 & 880 & 176 \\
        & BatteryDegradation1 & 256 & 1 & 366 & 153 \\
        \bottomrule
    \end{tabular}
\end{table}

\subsection{Preprocessing and Hyperparameters} \label{sec:preproc}

All inputs are standardized using training-set statistics. Regression targets are scaled to have unit mean for training and rescaled when reporting Root Mean Square Error (RMSE). Regularization weights $(\lambda_{\mathrm{spars}}, \lambda_{\mathrm{ortho}})$ are model-specific and reported in the corresponding experimental tables.

To ensure a fair comparison, all models share the same training protocol: 100 epochs, batch size 8, Adam optimizer ($\beta_1 = 0.9$, $\beta_2 = 0.999$), initial learning rate $10^{-3}$, and cosine annealing. No dataset-specific hyperparameter tuning is performed.

\textbf{MAGNETS architecture.}\quad The masking network is a 1D U-Net with three encoder and three decoder blocks (two convolutions per block, kernel size 3, ReLU), max-pooling in the encoder, and transposed convolutions in the decoder. Channel sizes are $\{32, 64, 128\}$ in the encoder and symmetrically reversed in the decoder. A final $1\times 1$ convolution outputs $(C\times M)$ logits per time step. We use $K=3$ concepts and $M=10$ masks for all experiments unless otherwise stated.

\textcolor{black}{To demonstrate the robustness of MAGNETS, we intentionally avoided per-dataset hyperparameter tuning, applying a single fixed configuration across all tasks. A sensitivity analysis (Appendix~\ref{sec:appendix_sensitivity}, Tables~\ref{tab:sensitivity_k} and~\ref{tab:sensitivity_lambda}) confirms that performance remains stable across a wide range of numbers of concepts and regularization weights.}

\textbf{Other baselines.}\quad OS uses the six summary features from \cite{wu_learning_2022}; NATM variants use multi-layer perceptrons (MLPs) with $32$ hidden units; GATSM uses the authors’ implementation; linear and tree-based models rely on \texttt{scikit-learn}; ROCKET/MultiROCKET use $10'000$ kernels from \texttt{sktime}. \textcolor{black}{Exact details about the CNN architectures (CNN, ResNet, InceptionTime) are given in Appendix~\ref{sec:appendix_hyperparameters}}.

\section{Results}\label{sec:results}

\subsection{Results on Synthetic Datasets}

\begin{table*}
    \centering
    \caption{Regression performance on the synthetic datasets. We report Root Mean Square Error (RMSE) ($\downarrow$) and $R^2$ ($\uparrow$) for black-box and interpretable baselines. Best results within each category (black-box or interpretable) are shown in \textbf{bold}, and the best overall result is \underline{\textbf{underlined}}. MAGNETS achieves top performance on three of the four datasets among interpretable models and competes closely with the strongest black-box baseline.}
    \label{tab:results_synth}
    \setlength{\tabcolsep}{2pt}
    \footnotesize
    \begin{tabular}{l >{\raggedright\arraybackslash}p{3cm} cccccccc}
        \toprule
        & & \multicolumn{2}{c}{Univariate} & \multicolumn{2}{c}{Bivariate} & \multicolumn{2}{c}{Trivariate-1} & \multicolumn{2}{c}{Trivariate-2}  \\
        & Model & RMSE $\downarrow$ & $R^2 \uparrow$ & RMSE $\downarrow$ & $R^2 \uparrow$ & RMSE $\downarrow$ & $R^2 \uparrow$ & RMSE $\downarrow$ & $R^2 \uparrow$ \\
        \midrule
        \multirow{6}{*}{\rotatebox[origin=c]{90}{Black-box}} & Random Forest & .2636 & .9383 & .5492 & .7155 & .5411 & .7380 & .5574 & .7221 \\[1pt]
        & ROCKET & .4458 & .8025 & .6639 & .5469 & .5152 & .7326 & .5343 & .7103 \\[1pt]
        & MultiROCKET & .1592 & .9748 & .6155 & .6112 & .4707 & .7768 & .4719 & .7739 \\[1pt]
        & \textcolor{black}{ResNet} & .0705 & .9949 & .1057 & .9879 & .0584 & .9965 & .0683 & .9952 \\
        & \textcolor{black}{InceptionTime} & .0475 & .9975 & .0950 & .9903 & .0583 & .9965 & .0671 & .9953 \\
        & CNN & \textbf{.0122} & \textbf{.9999} & \textbf{.0373} & \textbf{.9986} & \underline{\textbf{.0244}} & \underline{\textbf{.9994}} & \textbf{.0562} & \textbf{.9968} \\
        \midrule
        \multirow{11}{*}[-3.2em]{\rotatebox[origin=c]{90}{Interpretable}} & Mean & 1.003 & .0000 & .9863 & -.0003 & .9961 & .0000 & .9924 & .0000 \\ 
        & Linear regression & .6927 & .5229 & .9014 & .1645 & .5948 & .6434 & .6290 & .5983 \\
        & Lasso & .6925 & .5231 & .8991 & .1689 & .5941 & .6443 & .6280 & .5996 \\
        & Ridge & .6917 & .5243 & .8988 & .1694 & .5936 & .6450 & .6267 & .6012 \\
        \cmidrule{2-10}
        & Optimal Summaries [$\lambda_{\text{spars}}=\lambda_{\text{cos}}=0$] & .0976 & .9905 & .8355 & .2822 & .5378 & .7085 & .5748 & .6645 \\
        & Optimal Summaries [$\lambda_{\text{spars}}=\lambda_{\text{cos}}=0.1$] & .4395 & .8080 & .9863 & .0000 & .9965 & .0000 & .9923 & .0000 \\
        & NATM & .0510 & .9974 & .9077 & .1528 & .6011 & .6359 & .6336 & .5924 \\
        & NATM-Time & .0398 & .9984 & .9023 & .1629 & .5930 & .6456 & .6267 & .6012 \\
        & NATM-Feature & .0517 & .9973 & .9450 & .0818 & .8820 & .2160 & 1.276 & -.6538 \\
        & GATSM & \underline{\textbf{.0081}} & \underline{\textbf{.9999}} & .0901 & .9917 & .2131 & .9542 & .2196 & .9510 \\
        \cmidrule{2-10}
        & MAGNETS (Ours) [$\lambda_{\text{spars}},\lambda_{\text{ortho}}=0$] & .0273 & .9993 & \underline{\textbf{.0290}} & \underline{\textbf{.9991}} & \textbf{.0265} & \textbf{.9993} & \underline{\textbf{.0396}} & \underline{\textbf{.9984}} \\
        & MAGNETS (Ours) [$\lambda_{\text{spars}},\lambda_{\text{ortho}}=1$] & .0332 & .9989 & .0309 & .9990 & .0272 & \textbf{.9993} & .0455 & .9979 \\
        \midrule\midrule
        & MAGNETS vs. best BB & $+172\%$ & $-0.10\%$ & $-17.1\%$ & $+0.04\%$ & $+11.47\%$ & $-0.01\%$ & $-19.04\%$ & $+0.11\%$ \\
        & MAGNETS vs. best Int. & $+309\%$ & $+0.10\%$ & $-65.7\%$ & $+0.73\%$ & $-87.24\%$ & $+4.73\%$ & $-79.28\%$ & $+4.93\%$ \\
        \bottomrule
    \end{tabular}
\end{table*}

Table \ref{tab:results_synth} reports predictive performance on the four synthetic datasets. Although one might expect black-box models to dominate, the results are more nuanced: the CNN baseline achieves the lowest RMSE on the Univariate and Trivariate-1 datasets, whereas MAGNETS outperforms all black-box models\textemdash often by a substantial margin\textemdash on the Bivariate and Trivariate-2 tasks. Notably, MAGNETS achieves this level of accuracy while maintaining a fully transparent, end-to-end interpretable architecture.

Among interpretable baselines, MAGNETS is consistently the strongest performer. It clearly surpasses OS, all NATM variants, and GATSM on the Bivariate, Trivariate-1, and Trivariate-2 datasets, demonstrating its ability to capture nonlinear and multivariate temporal interactions. Only on the simplest Univariate task does GATSM perform better.

Overall, these results show that MAGNETS not only matches or exceeds the best black-box baselines on half of the tasks, but also outperforms every existing interpretable model across all multivariate settings.

\subsection{Interpretability Analysis on Synthetic Datasets} \label{sec:synth_interpretability}

\begin{figure}
    \centering
    \begin{subfigure}[b]{0.6\linewidth}
        \includegraphics[width=\textwidth]{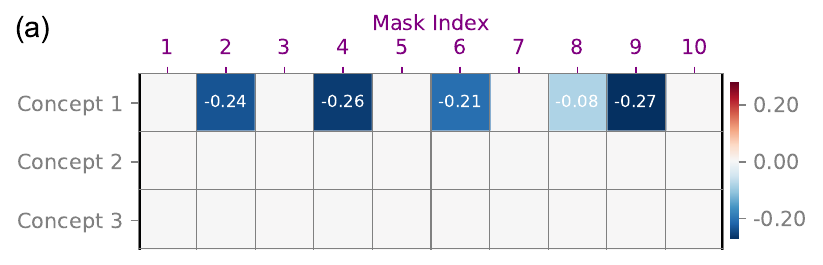}
        \label{fig:bottleneck-univariate:a}
    \end{subfigure}
    \begin{subfigure}[b]{0.6\linewidth}
        \vspace{-0.45cm}
        \includegraphics[width=\textwidth]{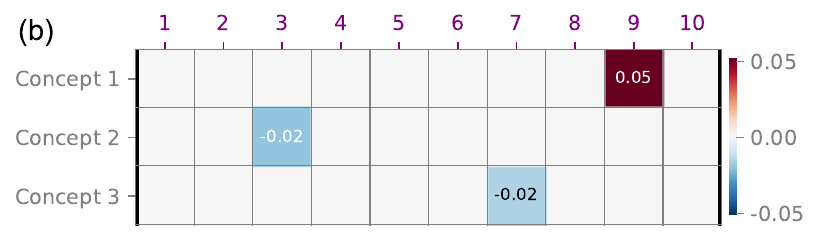}
        \label{fig:bottleneck-univariate:b}
        \vspace{-0.45cm}
    \end{subfigure}
    \begin{subfigure}[b]{0.6\linewidth}
        \includegraphics[width=\textwidth]{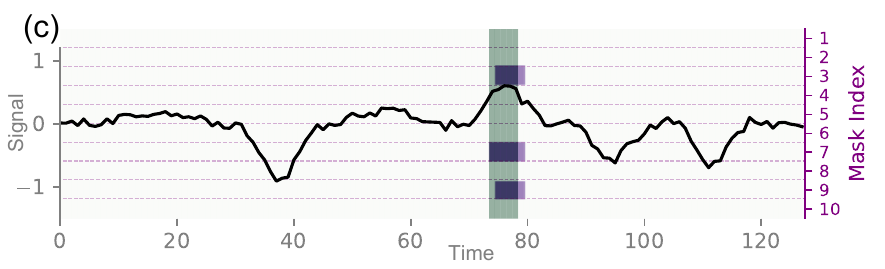}
        \label{fig:bottleneck-univariate-c}
        \vspace{-0.45cm}
    \end{subfigure}
    \begin{subfigure}[b]{0.6\linewidth}
        \includegraphics[width=\textwidth]{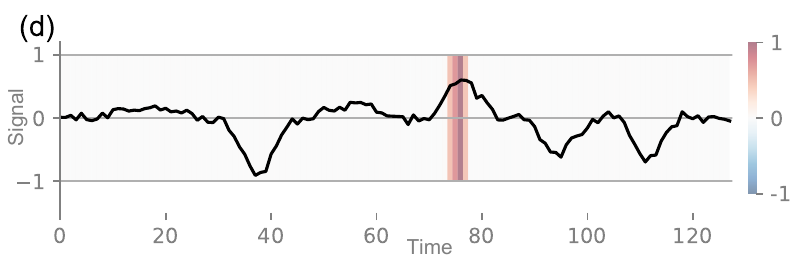}
        \label{fig:bottleneck-univariate:d}
        \vspace{-0.45cm}
    \end{subfigure}
    \caption{Concept bottleneck weights for a representative sample from the Univariate dataset. (a) Concepts learned without sparsity or orthogonality ($\lambda_{\mathrm{spars}},\lambda_{\mathrm{ortho}}=0$). (b) Concepts learned with both regularizers enabled ($\lambda_{\mathrm{spars}},\lambda_{\mathrm{ortho}}=1$). \linebreak(c) Mapping from learned masks to the input signal. Ground-truth intervals (shaded green) align with MAGNETS' identified regions (solid purple), confirming accurate recovery of the temporal logic. (d) DeepLift attributions for the black-box CNN.}
    \label{fig:bottleneck-univariate}
\end{figure}

\begin{figure*}
    \centering
    \begin{subfigure}[b]{1.0\linewidth}
        \includegraphics[width=\textwidth]{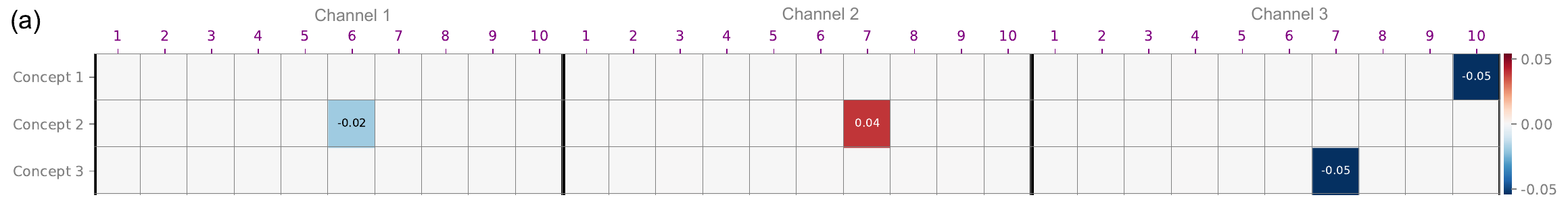}
        \label{fig:bottleneck-trivariate-2:a}
        \vspace{-0.45cm}
    \end{subfigure}
    \begin{subfigure}[b]{1.0\linewidth}
        \includegraphics[width=\textwidth]{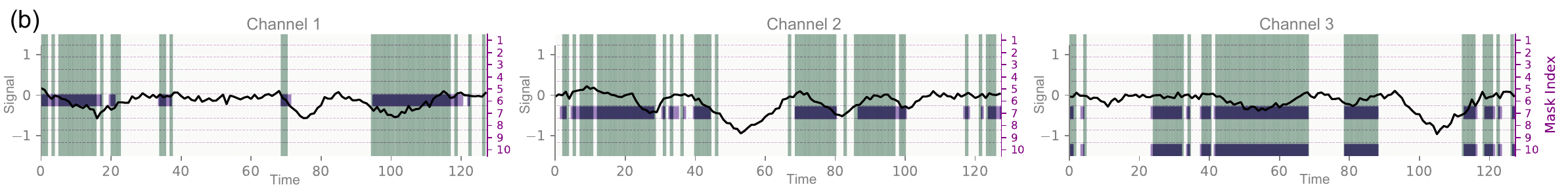}
        \label{fig:bottleneck-trivariate-2-b}
        \vspace{-0.45cm}
    \end{subfigure}
    \begin{subfigure}[b]{1.0\linewidth}
        \includegraphics[width=\textwidth]{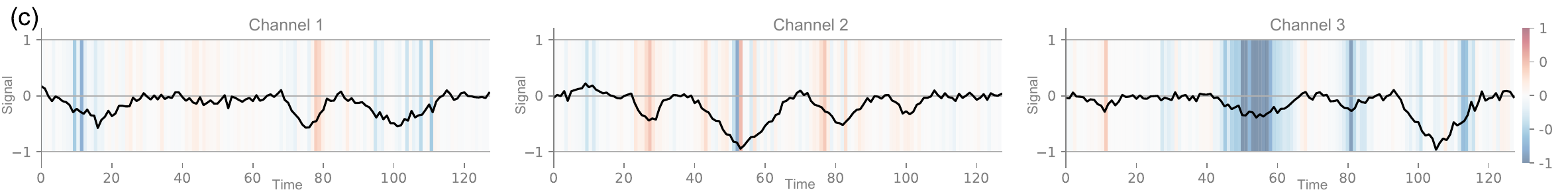}
        \label{fig:bottleneck-trivariate-2:c}
        \vspace{-0.45cm}
    \end{subfigure}
    \caption{Concept bottleneck weights for a representative sample from the Trivariate-2 dataset. (a) Concepts learned by MAGNETS ($\lambda_{\mathrm{spars}},\lambda_{\mathrm{ortho}}=1$). (b) Mapping from learned masks to the input signal, with ground-truth regions in green. Dim $i$ corresponds to the channel $i$ of the input time series. \textcolor{black}{Ground-truth intervals (shaded green) align with MAGNETS' identified regions (solid purple), confirming accurate recovery of the temporal logic.} (c) DeepLift attributions for the black-box CNN.}
    \label{fig:bottleneck-trivariate-2}
\end{figure*}

We now examine the interpretability of MAGNETS on the synthetic datasets by comparing the learned concepts against the known ground-truth rules. \textcolor{black}{Since TSER targets a single global scalar $\hat{y}$, there is no continuous target curve to plot. Instead, we highlight the exact ground-truth temporal intervals that mathematically drive the target in green shaded regions, shown on Fig.~\ref{fig:bottleneck-univariate}c and \ref{fig:bottleneck-trivariate-2}b.}

We focus on two representative cases: (i) the Univariate dataset, which illustrates how MAGNETS captures simple single-channel logic, and (ii) the Trivariate-2 dataset, which contains the richest and most complex multivariate interactions. Figures \ref{fig:bottleneck-univariate} and \ref{fig:bottleneck-trivariate-2} show the learned bottleneck weights as heatmaps (positive values in red, negative in blue), together with their corresponding mask activations over the raw input and the  ground-truth temporal regions they aim to recover.

\textbf{Univariate.}\quad Without regularization (Fig. \ref{fig:bottleneck-univariate}a), the concept bottleneck exhibits redundant structure: multiple concepts carry similar non-zero weights and activate overlapping masks. This redundancy is expected, as nothing in the unconstrained model prevents feature duplication  or compensation through the final linear layer. With regularization (Fig. \ref{fig:bottleneck-univariate}b), the bottleneck becomes sparse and disentangled, with each concept relying on a single dominant channel–mask pair. This structure reflects the simplicity of the underlying rule (“area above threshold”) and induces a clean one-to-one correspondence between concepts and meaningful temporal components. Fig. \ref{fig:bottleneck-univariate}c further shows that the resulting mask activations closely match the ground-truth region.

For comparison, Fig. \ref{fig:bottleneck-univariate}d presents DeepLift \citep{shrikumar_learning_2019} attributions for the black-box CNN. Although the CNN’s saliency map roughly highlights the correct interval, consistent with its low RMSE, the attributions fluctuate substantially across individual time steps rather than uniformly capturing the relevant region, illustrating the typical noisiness of post-hoc explanations.

\textbf{Trivariate-2.}\quad The Trivariate-2 task is the most challenging synthetic setting, requiring the model to combine three conditional areas defined by pairwise channel comparisons (Fig.~\ref{fig:synth})\textemdash a multivariate logic substantially more complex than simple thresholding. Despite the absence of any concept annotations, MAGNETS successfully recovers this structure. As shown in Fig.~\ref{fig:bottleneck-trivariate-2}a, the regularized bottleneck ($\lambda_{\mathrm{spars}}=\lambda_{\mathrm{ortho}}=1$) learns three disentangled concepts that correspond directly to the correct channel relationships. In particular, one concept depends jointly on channels 1 and 2, reflecting the interaction term in the true rule. The associated mask activations in Fig.~\ref{fig:bottleneck-trivariate-2}b accurately isolate the temporal regions used in the data-generation process, demonstrating that MAGNETS can reconstruct multi-channel symbolic logic directly from raw time series.

In contrast, the black-box CNN’s DeepLift attributions  (Fig.~\ref{fig:bottleneck-trivariate-2}c) highlight scattered individual time steps with inconsistent sign patterns (e.g. around timestep 50 on channel 2), making it difficult to reconstruct the underlying logic. In addition to exhibiting a higher RMSE than MAGNETS, its post-hoc explanations fail to capture the necessary temporal or multivariate relations. MAGNETS, by contrast, provides a clear and structured decomposition of \textit{which} channels interact, \textit{when} they matter, and \textit{how} they contribute to the prediction. This structured interpretability is achieved without sacrificing predictive performance, underscoring one of the main benefits of the proposed MAGNETS architecture.

\subsection{Results on Real-world Datasets}

\begin{table*}
    \vspace{-0.1cm}
    \caption{Regression performance on the real-world datasets. We report Root Mean Square Error (RMSE) ($\downarrow$) and $R^2$ ($\uparrow$) for black-box and interpretable baselines. Best results within each category (black-box or interpretable) are shown in \textbf{bold}, and the best overall result is \underline{\textbf{underlined}}. Linear regression fails to converge on HouseholdPowerC1 due to extreme outliers.}
    \centering
    \label{tab:results_real}
    \setlength{\tabcolsep}{3pt} 
    \footnotesize

    \hspace*{-1.5cm}
    \begin{tabular}{l >{\raggedright\arraybackslash}p{3cm} cccccccccc}
        \toprule
        & & \multicolumn{2}{c}{\makecell{Household\\PowerC1}} & \multicolumn{2}{c}{\makecell{WindTurbine\\Power}} & \multicolumn{2}{c}{\makecell{Bridge\\Degradation}}  & \multicolumn{2}{c}{\makecell{Benzene\\Concentration}} & \multicolumn{2}{c}{\makecell{\textcolor{black}{FloodModeling1}}} \\
        & Model & RMSE $\downarrow$ & $R^2 \uparrow$ & RMSE $\downarrow$ & $R^2 \uparrow$ & RMSE $\downarrow$ & $R^2 \uparrow$ & RMSE $\downarrow$ & $R^2 \uparrow$ & RMSE $\downarrow$ & $R^2 \uparrow$ \\
        \midrule
        \multirow{6}{*}{\rotatebox[origin=c]{90}{Black-box}} & Random Forest & 240.44 & .7832 & 107.72 & .9857 & 0.2323 & .9351 & 0.8480 & .9888 & .0360 & 0.2950 \\[1pt] 
        & ROCKET &  146.56 &  .9195 & 544.60 & .6353 & 0.2597 & .9248 & 21.972 & -6.516 & .0321 & .4371 \\[1pt] 
        & MultiROCKET & 152.15 & .9153 & 99.62 & .9878 & 0.1722 & .9724 & 10.995 & .4240 & \underline{\textbf{.0091}} & \underline{\textbf{.9548}} \\[1pt] 
        & ResNet & 136.46 & .8919 & 72.07 & .9911 & 0.0518 & .9957 & 1.222 & .9976 & .0232 & .6783 \\
        & InceptionTime & \underline{\textbf{85.058}} & .9103 & 65.734 & .9945 & 0.0485 & .9961 & 0.5985 & .9913 & .0100 & .9396 \\
        & CNN & 142.40 & \underline{\textbf{.9240}} & \underline{\textbf{23.41}} & \underline{\textbf{.9994}} & \underline{\textbf{0.0387}} & \underline{\textbf{.9978}} & \underline{\textbf{0.3994}} & \underline{\textbf{.9984}} & .0181 & .7989 \\ 
        \midrule
        \multirow{11}{*}[-3.2em]{\rotatebox[origin=c]{90}{Interpretable}} & Mean & 521.12 & -.0183 & 905.44 & -.0081 & 0.8410 & .0000 & 8.0152 & -.0001 & .0429 & -0.002 \\ 
        & Linear regression & NA & NA & 714.62 & .3720 & 0.3039 & .8694 & 1.6117 & .9596 & .2698 & -38.67 \\ 
        & Lasso & \textbf{151.99} & \textbf{.9134} & 448.28 & .7529 & 0.2405 & .9273 & 1.9793 & .9390 & .0414 & .0652 \\ 
        & Ridge & 152.70 & .9126 & 452.83 & .7478 & 0.2319 & .9310 & 1.6371 & .9583 & .0411 & .0805 \\ 
        \cmidrule{2-12}
        & Optimal Summaries [$\lambda_{\text{spars}},\lambda_{\text{cos}}=0$] & 181.40 & .8766 & 99.87 & .9877 & 0.3034 & .8637 & 7.9170 & .0242 & .0209 & .5775 \\
        & Optimal Summaries [$\lambda_{\text{spars}},\lambda_{\text{cos}}=0.1$] & 267.61 & .7314 & 551.67 & .6258 & 0.2968 & .8696 & 7.9710 & .0109 & .0198 & .5699 \\
        & NATM & 623.81 & -.4593 & 176.4 & .9617 & 0.0765 & .9913 & 4.9949 & .6116 & .5026 & -180.8 \\
        & NATM-Time & 167.33 & .8950 & 35.86 & .9984 & 0.1216 & .9781 & 57.407 & -50.30 & .1209 & -7.848 \\
        & NATM-Feature & 446.43 & .2523 & 175.83 & .9620 & 0.2826 & .8818 & 11.896 & -1.203 & .5063 & -185.8 \\
        & GATSM & 160.18 & .9038 & 355.47 & .8446 & \textbf{0.0326} & \textbf{.9984} & 1.5894 & .9607 & .0222 & .7001 \\
        \cmidrule{2-12}
        & MAGNETS (Ours) [$\lambda_{\text{spars}},\lambda_{\text{ortho}}=0$] & 153.20 & .9120 & \textbf{23.65} & \textbf{.9993} & 0.0670 & .9934 & 1.3262 & .9726 & .0187 & \textbf{.7936} \\ 
        & MAGNETS (Ours) [$\lambda_{\text{spars}},\lambda_{\text{ortho}}=1$] & 153.24 & .9119 & 24.63 & \textbf{.9993} & 0.0578 & .9957 & \textbf{1.2008} & \textbf{.9776} & \textbf{.0186} & .7911 \\ 
        \midrule\midrule
        \multicolumn{2}{l}{MAGNETS vs. best BB} & $+7.61\%$ & $-1.31\%$ & $+5.21\%$ & $-0.01\%$ & $+49.35\%$ & $-0.21\%$ & $+200\%$ & $-2.08\%$ & $+104\%$ & $-16.88\%$ \\
        \multicolumn{2}{l}{MAGNETS vs. best Int.} & $+0.82\%$ & $+0.16\%$ & $-31.31\%$ & $+0.10\%$ & $+77.30\%$ & $-0.27\%$ & $-24.45\%$ & $+1.76\%$ & $-6.06\%$ & $+13.36\%$ \\
        \bottomrule
    \end{tabular}

    \vspace{1em} 

    \hspace*{-1.5cm}
    \begin{tabular}{l >{\raggedright\arraybackslash}p{3cm} cccccccccc}
        \toprule
        & & \multicolumn{2}{c}{\textcolor{black}{\makecell{FloodModeling2}}} & \multicolumn{2}{c}{\makecell{\textcolor{black}{FloodModeling3}}}  & \multicolumn{2}{c}{\makecell{\textcolor{black}{BeijingPM10}\\\textcolor{black}{Quality}}} & \multicolumn{2}{c}{\makecell{\textcolor{black}{BeijingPM25}\\\textcolor{black}{Quality}}} & \multicolumn{2}{c}{\makecell{\textcolor{black}{Battery}\\\textcolor{black}{Degradation1}}} \\ 
        \midrule
        \multirow{6}{*}{\rotatebox[origin=c]{90}{Black-box}} & Random Forest & .0616 & .4046 & .0507 & .1939 & .4661 & .5065 & .4343 & .6875 & .0174 & .9420 \\[1pt] 
        & ROCKET & .0509 & .5943 & .0490 & .2471 & .6245 & .1142 & .6380 & .3254 & .0126 & .9697 \\[1pt] 
        & MultiROCKET & .0187 & .9452 & .0131 & .9463 & .5385 & .3412 & .5062 & .5754 & .0146 & .9596 \\[1pt] 
        & ResNet & .0294 & .8375 & .0339 & .6066 & .3766 & .5581 & .3394 & .7381 & .0580 & .1939 \\
        & InceptionTime & \textbf{.0116} & \textbf{.9565} & \underline{\textbf{.0107}} & \underline{\textbf{.9607}} & .3864 & \underline{\textbf{.5604}} & .3518 & .7378 & .0566 & .3285 \\
        & CNN & .0251 & .7041 & .0198 & .8475 & \underline{\textbf{.3634}} & .5588 & \textbf{.3191} & \textbf{.7510} & \textbf{.0117} & \textbf{.9704} \\ 
        \midrule
        \multirow{11}{*}[-3.2em]{\rotatebox[origin=c]{90}{Interpretable}} & Mean & .0800 & -0.0042 & .0568 & -0.011 & .6694 & -0.018 & .7789 & -0.005 & .0728 & -0.012 \\ 
        & Linear regression & .1512 & -2.583 & 1.352 & -572.1 & .4872 & .4608 & .4787 & .6203 & .0180 & .9377 \\ 
        & Lasso & .0825 & -0.0667 & .0550 & .0498 & .4857 & .4641 & .4747 & .6266 & .0361 & .7510 \\ 
        & Ridge & .0824 & -0.0649 & .0551 & .0486 & .4864 & .4626 & .4764 & .6239 & .0217 & .9102 \\ 
        \cmidrule{2-12}
        & Optimal Summaries [$\lambda_{\text{spars}},\lambda_{\text{cos}}=0$] & .0262 & .7363 & .0269 & .7205 & .3802 & .4906 & .3696 & .6558 & .0289 & .7890 \\
        & Optimal Summaries [$\lambda_{\text{spars}},\lambda_{\text{cos}}=0.1$] & .0271 & .6438 & .0307 & .6509  & .3847 & .4820 & .3739 & .6440 & .0540 & .2406 \\
        & NATM & 1.044 & -641.6 & .7904 & -242.57 & .4078 & .4606 & .3978 & .6355 & .1186 & -3.056 \\
        & NATM-Time & 1.045 & -641.5 & .1263 & -4.527 & .4064 & .4265 & .3932 & .5954 & .6582 & -151.13 \\
        & NATM-Feature & 1.044 & -641.5 & .7904 & -242.57 & .4708 & .3265 & .4706 & .5171 & .0559 & .2023 \\
        & GATSM & .0500 & -0.098 & .0174 & .8822 & \textbf{.3729} & \textbf{.5202} & .3683 & .6368 & .0328 & .7797 \\
        \cmidrule{2-12}
        & MAGNETS (Ours) [$\lambda_{\text{spars}},\lambda_{\text{ortho}}=0$] & \underline{\textbf{.0084}} & \underline{\textbf{.9837}} & .0171 & .8983 & .3826 & .4705 & .3579 & \textbf{.6677} & \underline{\textbf{.0115}} &  \underline{\textbf{.9708}} \\
        & MAGNETS (Ours) [$\lambda_{\text{spars}},\lambda_{\text{ortho}}=1$] & .0184 & .8912 & \textbf{.0170} & \textbf{.9030} & .3933 & .4539 & \textbf{.3571} & .6531 & .0216 & .8808 \\ 
        \midrule\midrule
        \multicolumn{2}{l}{MAGNETS vs. best BB} & $-27.59\%$ & $+2.84\%$ & $+58.88\%$ & $-6.00\%$ & $+5.28\%$ & $-16.04\%$ & $+11.91\%$ & $-11.09\%$ & $-1.71\%$ & $+0.04\%$ \\
        \multicolumn{2}{l}{MAGNETS vs. best Int.} & $-67.94\%$ & $+33.60\%$ & $-36.80\%$ & $+2.36\%$ & $+2.60\%$ & $-9.55\%$ & $-3.38\%$ & $+1.81\%$ & $-60.21\%$ & $+23.04\%$ \\
        \bottomrule
    \end{tabular}
\end{table*}

Table~\ref{tab:results_real} reports the predictive performance across all ten real-world TSER datasets. As expected, unconstrained black-box models generally achieve the lowest RMSE due to the performance-interpretability trade-off. Nevertheless, MAGNETS narrows this gap and remains highly competitive, even outperforming all baselines on FloodModeling2 and BatteryDegradation1. Additionally, its accuracy closely approaches the best black-box model on HouseholdPowerC1 and WindTurbinePower, and remains within a narrow margin on BeijingPM10Quality, BeijingPM25Quality, and BridgeDegradation, while preserving an inherently interpretable computation path.

Among interpretable models, MAGNETS is the consistently strongest performer. It achieves the lowest RMSE on seven of the ten datasets (WindTurbinePower, BenzeneConcentration, FloodModeling1, FloodModeling2, FloodModeling3, BeijingPM25Quality, and BatteryDegradation1). On the remaining tasks, it remains highly competitive, often nearly tying with the top transparent baseline (\textit{e.g.}, Lasso on HouseholdPowerC1 and GATSM on BeijingPM10Quality). Notably, no other interpretable method achieves competitive performance across all ten datasets, highlighting the robustness and versatility of MAGNETS.

\subsection{Evaluation of Explanation Quality}

We quantitatively compare MAGNETS’ learned masks to the ground-truth (GT) temporal regions and to post-hoc attribution maps produced by Integrated Gradients (IG) and DeepLIFT (DL) applied to the best black-box model (CNN). For each synthetic dataset, GT masks are defined as binary indicators marking the time steps that contribute to the target. Because the evaluation focuses on \textit{when} relevant regions occur (not \textit{why}), masks are aggregated across channels by taking a per-time-step maximum. Only samples with non-empty GT masks are considered.

IG and DL attributions are normalized to $[-1,1]$, mapped to $[0,1]$ by taking absolute values, and compared to GT using AUC (on continuous masks) and F1 score (after thresholding at 0.5). MAGNETS' masks are evaluated with the same procedure.

As shown in Table \ref{tab:explanation-quality}, IG and DL achieve high AUC on the simpler datasets (Univariate and Bivariate) but their performance deteriorates substantially as multivariate interactions become more complex, dropping to only $0.66$ AUC on Trivariate-2. In contrast, MAGNETS maintains consistently high AUC and F1 scores across all datasets, with only a modest decrease on the most challenging task. These results indicate that MAGNETS provides robust and faithful explanations even in settings where post-hoc attribution methods fail to capture the underlying temporal logic.

\begin{table}[ht]
    \centering
    \caption{Explanation quality on the synthetic datasets. We report AUC ($\uparrow$) and F1 scores ($\uparrow$) comparing MAGNETS masks and post-hoc attribution baselines to the ground-truth regions. Best results are shown in \textbf{bold}.}
    \label{tab:explanation-quality}
    \setlength{\tabcolsep}{2pt}
    \small
    \begin{tabular}{lcccccccc}
        \toprule
        & \multicolumn{2}{c}{Univariate} & \multicolumn{2}{c}{Bivariate} & \multicolumn{2}{c}{Trivariate-1} & \multicolumn{2}{c}{Trivariate-2}  \\
        Method & AUC & F1 & AUC & F1 & AUC & F1 & AUC & F1 \\
        \midrule
        CNN + Integrated Gradients & 0.99 & 0.69 & 0.98 & 0.49 & 0.83 & 0.36 & 0.66 & 0.34 \\
        CNN + DeepLIFT & 0.99 & 0.75 & 0.95 & 0.48 & 0.77 & 0.35 & 0.66 & 0.34 \\
        \midrule
        MAGNETS [$\lambda_{\text{spars}},\lambda_{\text{ortho}}=0$] & \textbf{1.00} & \textbf{1.00} & \textbf{1.00} & \textbf{1.00} & \textbf{1.00} & \textbf{1.00} & 0.98 & 0.94 \\
        MAGNETS [$\lambda_{\text{spars}},\lambda_{\text{ortho}}=1$] & 0.99 & 0.93 & \textbf{1.00} & \textbf{1.00} & \textbf{1.00} & \textbf{1.00} & \textbf{0.99} & \textbf{0.95} \\
        \bottomrule
    \end{tabular}
\end{table}

\subsection{Interpretability Analysis on Real-world Datasets} \label{sec:real_interpretability}

\begin{figure*}
    \centering
    \begin{subfigure}[b]{1.0\linewidth}
        \includegraphics[width=\textwidth]{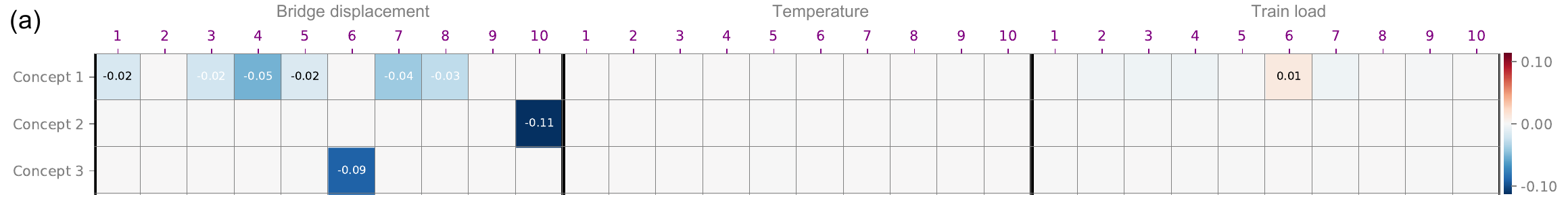}
        \label{fig:bottleneck-bridge:a}
        \vspace{-0.45cm}
    \end{subfigure}
    \begin{subfigure}[b]{1.0\linewidth}
        \includegraphics[width=\textwidth]{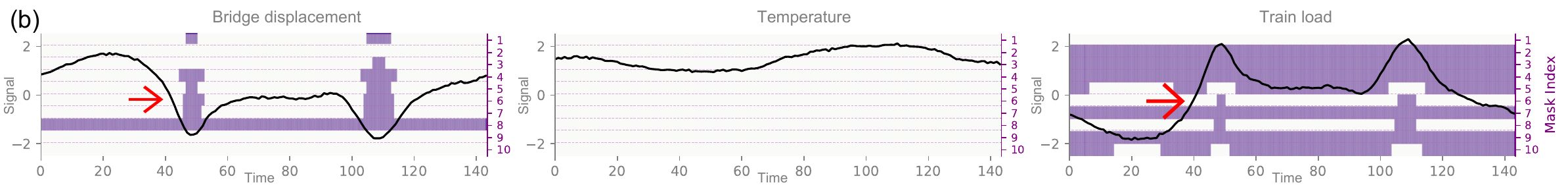}
        \label{fig:bottleneck-bridge:b}
        \vspace{-0.45cm}
    \end{subfigure}
    \begin{subfigure}[b]{1.0\linewidth}
        \includegraphics[width=\textwidth]{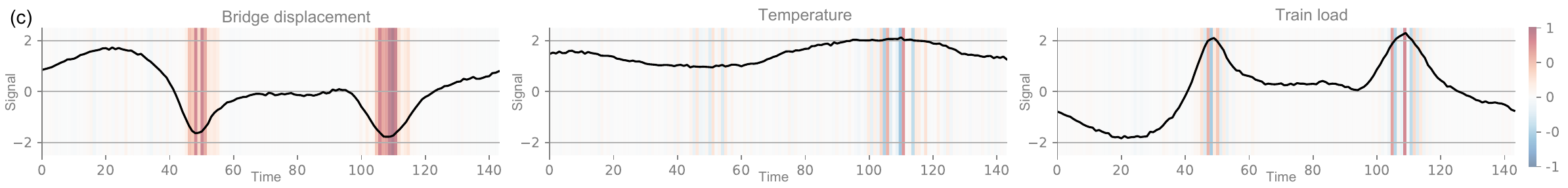}
        \label{fig:bottleneck-bridge:c}
        \vspace{-0.45cm}
    \end{subfigure}
    \caption{Concept bottleneck weights for a representative sample from the BridgeDegradation dataset. (a) Concepts learned by MAGNETS ($\lambda_{\mathrm{spars}},\lambda_{\mathrm{ortho}}=1$). (b) Mapping from learned masks to the input signal, with red arrows showing important masks. Dim $i$ corresponds to the channel $i$ of the input time series. (c) DeepLift attributions for the black-box CNN.}
    \label{fig:bottleneck-bridge}
\end{figure*}

We qualitatively assess the interpretability of MAGNETS on the BridgeDegradation dataset \textcolor{black}{and on the BatteryDegradation1 dataset (Appendix~\ref{sec:appendix_battery})}, which provides an illustrative real-world example without ground-truth concept annotations.

Figure~\ref{fig:bottleneck-bridge}a displays the learned bottleneck weights \textcolor{black}{on a BridgeDegradation sample}. Concept 1 assigns non-zero weights to masks on bridge displacement and train load, indicating that the model has discovered a meaningful multivariate dependency between these signals. Concept 3 strongly activates mask~6 on bridge displacement, whereas Concept 2 primarily attends to mask 10, which contains no active time steps (see Fig.~\ref{fig:bottleneck-bridge}b). This indicates that Concept~2 behaves effectively as an input-independent bias term rather than capturing a temporal pattern.

Figure~\ref{fig:bottleneck-bridge}b (red arrows) highlights the key masks associated with Concepts 1 and 3. Examining the corresponding masked regions reveals two consistent temporal patterns: (i) peaks in train load above approximately 2.0, and (ii) the corresponding values of bridge displacement at those same time steps. These observations support a plausible hypothesis for domain practitioners: the target bridge degradation $y$ depends primarily on the bridge displacement during intervals when train load exceeds a critical threshold.

For comparison, Figure~\ref{fig:bottleneck-bridge}c presents DeepLIFT attributions for the black-box CNN. Although the CNN attends to broadly similar intervals, its attributions are less structured: they fluctuate sharply across neighboring time steps, exhibit oscillating positive and negative contributions, and assign weak relevance to temperature despite its unclear role. Because these attributions are post-hoc, their fidelity to the model’s internal reasoning is uncertain. By contrast, MAGNETS provides inherently faithful explanations, as the identified masks directly drive the prediction.

\textcolor{black}{A parallel explainability analysis on the BatteryDegradation1 dataset is provided in Appendix~\ref{sec:appendix_battery}, demonstrating that MAGNETS learns and extract interpretable masks aligned with physical battery chemistry.}

Overall, MAGNETS yields stable, interpretable concepts that support clearer hypothesis formation and more actionable insights for practitioners.

\section{Conclusion and Future Work}\label{sec:conclusion}

In this work, we propose MAGNETS, an interpretable neural architecture for time series regression that learns input-dependent masks and aggregates them into meaningful concepts without requiring concept supervision. \textcolor{black}{Overall, MAGNETS directly addresses the performance-interpretability trade-off by narrowing the gap with black-box baselines while outperforming existing interpretable architectures on tasks aligned with localized temporal evidence.}

Experiments on synthetic and real-world datasets show that MAGNETS achieves accuracy competitive with strong black-box models while consistently outperforming existing interpretable baselines. \textcolor{black}{Furthermore, when the regression task strongly aligns with its localized inductive bias, MAGNETS can even surpass black-box performance}. On synthetic tasks with known generative structure, MAGNETS not only identifies the correct temporal regions but also reconstructs advanced multivariate dependencies without supervision, demonstrating the faithfulness and expressiveness of its explanations.

Several limitations open promising directions for future research. For real-world datasets, quantitative validation of explanation correctness remains limited by the absence of ground-truth temporal annotations. Additionally, the current aggregation mechanism is limited to summation, constraining the class of functions the model can represent. Exploring alternative differentiable aggregators could broaden its applicability. Expressiveness may also be enhanced through polynomial expansions or carefully designed nonlinearities, though such extensions must be balanced against interpretability. Finally, extending MAGNETS to time series classification, where temporal localization and concept-level reasoning are equally valuable, represents a natural next step.

\backmatter

\bmhead{Acknowledgements}

This work was supported by the Swiss National Science Foundation grant $200021\_200461$. We thank the teams at UCR, UEA, and Monash University for their efforts in providing the benchmark datasets.

\section*{Statements and Declarations}

\bmhead{Funding} This work was supported by the Swiss National Science Foundation grant $200021\_200461$.

\bmhead{Competing interests} The authors have no competing interests to declare that are relevant to the content of this article.

\bmhead{Data and code availability} The code implementation and datasets are publicly available at \url{https://github.com/FlorentF9/MAGNETS}.

\bmhead{Author contributions}

F.F., A.W., and O.F. conceptualized the idea, F.F. developed the methodology, F.F., A.W., and O.F. conceived the experiments, F.F. and A.W. conducted the experiments, F.F., A.W., and O.F. analyzed the results, and all authors wrote the manuscript.


\bibliography{sn-bibliography}

\appendix

\section{Hyperparameter Sensitivity Analysis} \label{sec:appendix_sensitivity}

\textcolor{black}{To validate that MAGNETS does not rely on per-dataset hyperparameter tuning, we conducted an extensive sensitivity analysis on the Trivariate-2 and BridgeDegradation dataset. We evaluated the model's performance across varying numbers of concepts ($K$) and a grid of regularization weights ($\lambda_{\mathrm{spars}}$ and $\lambda_{\mathrm{ortho}}$).}

\textcolor{black}{As shown in Tables \ref{tab:sensitivity_k} and \ref{tab:sensitivity_lambda}, even extreme deviations from the default configuration cause minor fluctuations. MAGNETS remains highly stable across the tested range on the controlled Trivariate-2 benchmark. On BridgeDegradation, performance is more sensitive, especially in \(R^2\), suggesting that real-world datasets may benefit from validation-based hyperparameter selection when maximum predictive accuracy is the primary objective. Nevertheless, the default configuration remains competitive without per-dataset tuning.}

\captionsetup[table]{labelfont={bf,color=black}}
\begin{table}[ht]
    \captionsetup{width=\textwidth} 
    \caption{\textcolor{black}{Sensitivity analysis of MAGNETS with respect to the number of concepts $K$. Regularization weights are fixed at $\lambda_\mathrm{spars} = 1.0$ and $\lambda_\mathrm{ortho} = 1.0$. The default configuration is $K = 3$.}}
    \label{tab:sensitivity_k}
    \centering
    \begin{tabular}{ccccc}
        \toprule
        & \multicolumn{2}{c}{\textbf{Trivariate-2}} & \multicolumn{2}{c}{\textbf{BridgeDegradation}} \\
        \cmidrule(lr){2-3} \cmidrule(lr){4-5}
        $\boldsymbol{K}$ & \textbf{RMSE} ($\downarrow$) & \textbf{$R^2$} ($\uparrow$) & \textbf{RMSE} ($\downarrow$) & \textbf{$R^2$} ($\uparrow$) \\
        \midrule
        3           & .0455 & .9979 & .0578 & .9957 \\
        5           & .0450 & .9964 & .0573 & .9120 \\
        10          & .0421 & .9968 & .0546 & .9036 \\
        15          & .0502 & .9957 & .0622 & .8750 \\
        \bottomrule
    \end{tabular}
\end{table}

\captionsetup[table]{labelfont={bf,color=black}}
\begin{table}[ht]
\captionsetup{width=\textwidth} 
\centering
\caption{\textcolor{black}{\textbf{Sensitivity analysis of MAGNETS with respect to the regularization weights $\lambda_\mathrm{spars}$ and $\lambda_\mathrm{ortho}$}. The number of concepts is fixed at $K=3$. One weight is varied while the other is held at its default value of 1.0.}}
\label{tab:sensitivity_lambda}
\begin{tabular}{cccccc}
\toprule
 & & \multicolumn{2}{c}{\textbf{Trivariate-2}} & \multicolumn{2}{c}{\textbf{BridgeDegradation}} \\
\cmidrule(lr){3-4} \cmidrule(lr){5-6}
$\boldsymbol{\lambda_\mathrm{spars}}$ & $\boldsymbol{\lambda_\mathrm{ortho}}$ & \textbf{RMSE} ($\downarrow$) & $\boldsymbol{ R^2}$ ($\uparrow$) & \textbf{RMSE} ($\downarrow$) & $\boldsymbol{ R^2}$ ($\uparrow$) \\
\midrule
1.0          & 1.0          & .0455 & .9979 & .0578 & .9957 \\
\midrule
0.1          & 1.0          & .0449 & .9964 & .0519  & .8920 \\
0.5          & 1.0          & .0443 & .9965 & .0600  & .8733 \\
10.0         & 1.0          & .0505 & .9955 & .0798 & .8341 \\
\midrule
1.0          & 0.1          & .0399 & .9972 & .0672 & .8268 \\
1.0          & 0.5          & .0408 & .9971 & .0532 & .8943 \\
1.0          & 10.0         & .0431 & .9967 & .0493 & .9089 \\
\bottomrule
\end{tabular}
\end{table}

\newpage
\section{Baseline Hyperparameters} \label{sec:appendix_hyperparameters}

\textcolor{black}{The CNN baseline consisted of three 1D convolutional blocks with channel widths 32, 64, and 128 each using kernel size 3, ReLU activation, and max-pooling with stride 2. The resulting feature map was flattened and projected to a 128-dimensional latent representation, followed by a linear regression head.}

\textcolor{black}{ResNet consists of an initial 1D convolutional layer followed by a stack of 4 residual blocks with channel widths $[64, 128, 196, 256]$. InceptionTime consists of 6 inception modules with 32 filters per branch (kernel sizes 39, 19, and 9, followed by a max-pooling convolutional branch). For multivariate inputs, a 32-channel bottleneck was applied before the convolution branches. All models were trained for 100 epochs using Adam (learning rate $10^{-3}$) with a cosine annealing learning rate schedule.}

\newpage
\section{Additional Qualitative Analysis on Explainability}\label{sec:appendix_battery}

\begin{figure*}[h]
    \centering
    \includegraphics[width=\linewidth]{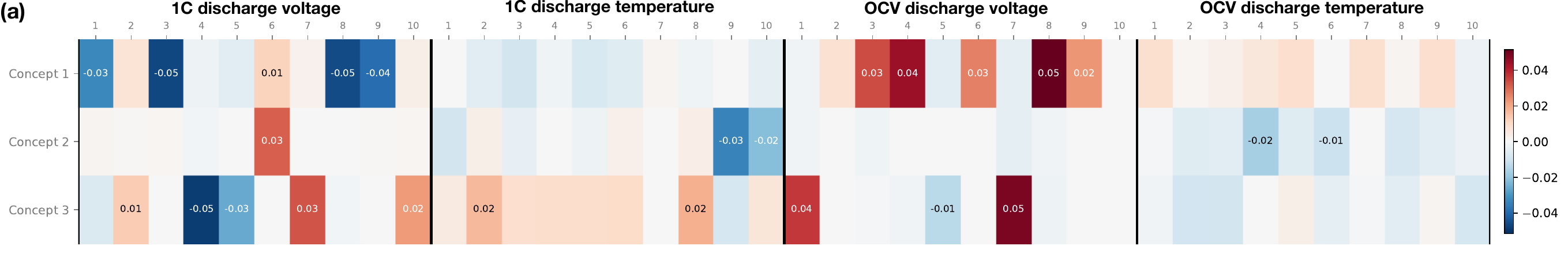}
    \includegraphics[width=\linewidth]{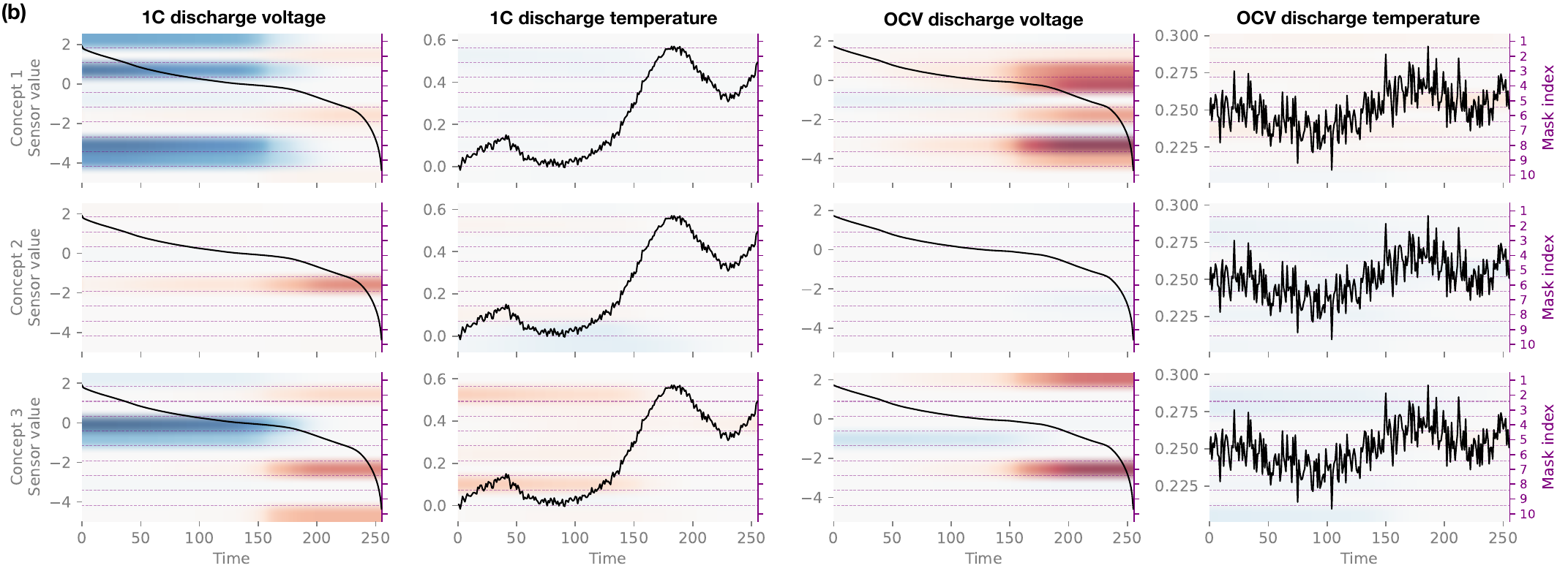}
    \vspace{-0.6cm}
    \caption{\textcolor{black}{\textbf{Extracted concepts by MAGNETS ($K=3$, $M=10$) on a sample from the OxfordBatteryDegradation1 dataset.} (a) Concept bottleneck layer with individual mask weights per input channel (four channels in total). Each individual column represents a mask. (b) Overlay of the raw input signals (black) with weighted mask activations by concept. Blue and red represent negative and positive weights, respectively. MAGNETS focuses on two parts of the time-series to predict the battery state-of-health: the beginning of the 1C discharge voltage, and the end of the OCV discharge voltage.}}
    \label{fig:battery_masks}
    \vspace{-0.6cm}
\end{figure*}

\textcolor{black}{To provide further interpretability, we qualitatively analyze the extracted masks by MAGNETS on a sample from the BatteryDegradation1 dataset (see Fig.~\ref{fig:battery_masks}).}

\textcolor{black}{The objective of this TSER task is to predict the battery's state-of-health using signals from two standard characterization protocols: a constant-current 1C discharge cycle and an Open Circuit Voltage (OCV) discharge cycle. As illustrated in Figure~\ref{fig:battery_masks}, MAGNETS successfully extracts concepts that isolate distinct temporal features across the four input channels. Specifically, the weighted mask activations overlaid on the raw signals (Fig.~\ref{fig:battery_masks}b) reveal that the model heavily focuses on the initial phase of the 1C discharge voltage and the final tail of the OCV discharge voltage.}

\textcolor{black}{This localized focus makes strong physical sense, as the initial voltage drop under a 1C load is a primary indicator of internal resistance growth, while the discharge profile at the end of an OCV cycle directly reflects the remaining active material capacity~\citep{birkl2017degradation}. Ultimately, this confirms that the representations automatically discovered by our method correspond to the underlying, physically meaningful degradation mechanisms.}

\captionsetup[figure]{labelfont={bf,color=black}}

\end{document}